\documentclass[10pt,twocolumn,letterpaper]{article}

\usepackage{iccv}
\usepackage{times}
\usepackage{epsfig}
\usepackage{graphicx}
\usepackage{amsmath}
\usepackage{amssymb}

\usepackage{algorithm}
\usepackage{algorithmic}
\usepackage{multirow}
\usepackage{arydshln}

\usepackage[pagebackref=true,breaklinks=true,letterpaper=true,colorlinks,bookmarks=false]{hyperref}

\iccvfinalcopy 

\def\iccvPaperID{10708} 

\ificcvfinal\fi

\begin{document}

\title{Coarse-to-Fine: Learning Compact Discriminative Representation for Single-Stage Image Retrieval}

\author{Yunquan Zhu\thanks{Equal contribution.}~,~~ Xinkai Gao$^*$,~~  Bo Ke,~~  Ruizhi Qiao,~~  Xing Sun\\
YouTu Lab, Tencent, China\\
{\tt\normalsize \{yunquanzhu, kayxgao, boke, ruizhiqiao, winfredsun\}@tencent.com}
}

\maketitle
\ificcvfinal\thispagestyle{empty}\fi

\begin{abstract}
   Image retrieval targets to find images from a database that are visually similar to the query image. Two-stage methods following retrieve-and-rerank paradigm have achieved excellent performance, but their separate local and global modules are inefficient to real-world applications. To better trade-off retrieval efficiency and accuracy, some approaches fuse global and local feature into a joint representation to perform single-stage image retrieval. However, they are still challenging due to various situations to tackle, $e.g.$, background, occlusion and viewpoint. In this work, we design a $\mathbf{C}$oarse-to-$\mathbf{F}$ine framework to learn $\mathbf{C}$ompact $\mathbf{D}$iscriminative representation (CFCD) for end-to-end single-stage image retrieval-requiring only image-level labels. Specifically, we first design a novel adaptive softmax-based loss which dynamically tunes its scale and margin within each mini-batch and increases them progressively to strengthen supervision during training and intra-class compactness. Furthermore, we propose a mechanism which attentively selects prominent local descriptors and infuse fine-grained semantic relations into the global representation by a hard negative sampling strategy to optimize inter-class distinctiveness at a global scale. Extensive experimental results have demonstrated the effectiveness of our method, which achieves state-of-the-art single-stage image retrieval performance on benchmarks such as Revisited Oxford and Revisited Paris. Code is available at \href{https://github.com/bassyess/CFCD}{https://github.com/bassyess/CFCD}.
\end{abstract}

\begin{figure}[t]
  \centering
  \includegraphics[width=1.0\linewidth]
                  {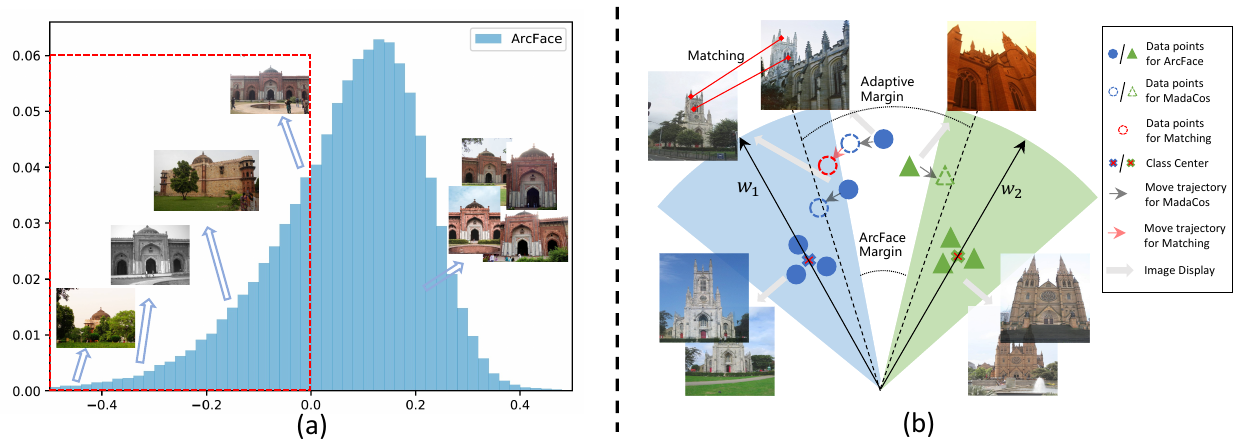}
  \caption{
    (a) Distribution of $[cos(\theta_{y_i}+m)-argmax(cos\theta_{j\ne y_i})]$ after training with ArcFace, where $y_i$ is target label. In the red box, due to large variations in background, occlusion and viewpoint, these images in Google Landmarks Dataset V2 are far away from their class centers and misclassified. (b) Geometrical interpretation of our methods from the feature perspective. 
    By designing an adaptive margin penalty strategy, we can introduce appropriate supervision intensity for different batch during training.
    As for the outliers with partial match, we design a mechanism to select prominent local descriptors and minimize their pairwise distances, which makes the unified representation more discriminative. 
  }
  \label{fig:motivation}
  \vspace{-1mm}
\end{figure}

\section{Introduction}

Image retrieval is a fundamental task in computer vision, which aims to efficiently retrieve images similar to a given query from a large-scale database. With the development of deep learning, image retrieval has made great progress~\cite{musgrave2020metric,tan2021instance, wu2022learning, deng2022insclr, chen2021deep}. The state-of-the-art methods generally work in a two-stage paradigm~\cite{cao2020unifying, lee2022cvnet}, where they first obtain coarse candidates via global features, and then re-rank them with local features to achieve better performance. However, two-stage methods are required to rank images twice and use the expensive RANSAC~\cite{fischler1981random} or AMSK~\cite{tolias2016image} for geometric verification, leading to high memory usage and increased latency.

To alleviate the efficiency issues, many studies~\cite{gordo2016deep, movshovitz2017no, kim2020proxy,el2021training} recently attempt to explore a unified single-stage image retrieval solution. They design complicated attention modules to fuse global and local features, and adopt the ArcFace~\cite{deng2019arcface} loss to train the model in an end-to-end fashion. They have shown excellent performance on single-stage image retrieval benchmarks. In spite of their successes, extracting multi-scale local features is still an extremely expensive process. More importantly, these studies do not consider the challenges of large-scale landmark dataset from the perspective of data distribution, which have large variations in background, occlusion and viewpoint.

Fig\ref{fig:motivation}(a) displays the cosine logits distribution of landmark samples after convergence, where more than 20\% of the samples are far away from their class centers as their target cosine logits are smaller than their non-target cosine logits. One can observe the various conditions in these samples such as background, occlusion, viewpoint, \etc. 
Moreover, true positives lingering at the classification boundary receive weaker supervision due to the fixed margin penalty. Therefore, we propose an adaptive margin penalty strategy that tunes hyper-parameters to progressively strengthen supervision for  intra-class compactness. Besides, inspired by geometric verification, in order to retrieve target images with partial match, we design a mechanism to select prominent local descriptors and minimize their pairwise distances for learning inter-class distinctiveness more effectively, is shown in Fig\ref{fig:motivation}(b).

We propose a $\mathbf{C}$oarse-to-$\mathbf{F}$ine framework to learn $\mathbf{C}$ompact and $\mathbf{D}$iscriminative representation (CFCD) for single-stage image retrieval.
Specifically, we first propose a novel adaptive loss which uses the median of cosine logits in a batch to dynamically tune the scale and margin of the loss function, namely MadaCos. MadaCos increases its scale and margin progressively to strengthen supervision during training, consequently increasing the learned intra-class compactness. We also design the local descriptors matching constraints and hard negative sampling strategy to construct triplets, and introduce the triplet loss\cite{bronstein2010data} to leverage fine-grained semantic relations, which embed the global feature with more information of inter-class distinctiveness. We jointly train the model as a whole with MadaCos and triplet losses to produce the final compact and discriminative representation and improve the overall performance. This framework consists of two training phases: global feature learning with MadaCos and later added local feature matching with triplet loss. During the testing stage, global features are extracted from the the end-to-end framework and ranked once without additional computation overhead. Our main contributions are summarized as follows:

\begin{itemize}
\item[$\bullet$] We propose a coarse-to-fine framework to learn compact and discriminative representation for single-stage image retrieval without additional re-ranking computation overhead, which is more efficient.
\item[$\bullet$] To enhance intra-class compactness, we design an adaptive softmax loss named MadaCos, which uses the median of cosine logits within each mini-batch to tune its hyperparameters to strengthen supervision. 
\item[$\bullet$] To enhance inter-class distinctiveness, we select prominent local descriptors and design an image-level hard negative sampling strategy to leverage fine-grained semantic relations.  
\item[$\bullet$] Through systematic experiments, the proposed method achieves stage-of-the-art single-stage image retrieval performance on benchmarks: $\mathcal{R}$Oxf (+1M), $\mathcal{R}$Par (+1M).
\end{itemize}

\section{Related Work}
\subsection{Image Retrieval}
In early researches, global features are developed by aggregating hand-crafted local features through Fisher vector~\cite{jegou2011aggregating}, VLAD~\cite{jegou2010aggregating} or ASMK~\cite{tolias2016image}.
Afterward, spatial verification performs local features matching with RANSAC~\cite{fischler1981random} to re-rank preliminary retrieval results, which effectively improves the overall performance. 
Recently, handcrafted features have been replaced by global and local features extracted from deep learning networks. 
In local features based image retrieval, \cite{balntas2016learning,zheng2017sift,he2018local,dusmanu2019d2,superfeatures} have made remarkable progress by leveraging discriminative geometry information. 
From the global aspect, high-level semantic features are obtained simply by performing differentiable pooling operations such as sum-pooling(SPoC)\cite{babenko2015aggregating}, regional-max-pooling(R-MAC)\cite{gordo2017end} and generalized mean-pooling(GeM)\cite{radenovic2018fine} on the feature maps of CNNs. 
The state-of-the-art approaches leverage local and global features to explore two types of algorithm pipelines. 
In the two-stage paradigm, the typical method DELG~\cite{cao2020unifying} incorporated with DELF’s~\cite{noh2017large} attention module trains local and global features in an end-to-end manner. They first search by global features, then re-rank the top database images using local feature matching. 
To alleviate the problems of high memory usage and latency of two-stage methods, the single-stage methods such as Token \cite{wu2022learning} and DOLG\cite{yang2021dolg} fuse local features or both local and global features into a compact representation, and only rank images once.
However, the two categories either suffer from inefficient local branches or weak supervisions of local features. Our work is essentially different from them. We propose a coarse-to-fine framework to quickly train coarse global and local descriptors, and then select matching local descriptors to refine the global features to integrate local fine-grained information. In other words, we introduce the idea of geometric validation to supervise local features in an end-to-end training scheme, which replaces the complicated local branches. Moreover, the inference of our framework is performed without additional re-ranking computation overhead.

\begin{figure*}[t]
  \centering
  \setlength{\abovecaptionskip}{-0.5mm}
  \includegraphics[width=1.0\linewidth]
                  {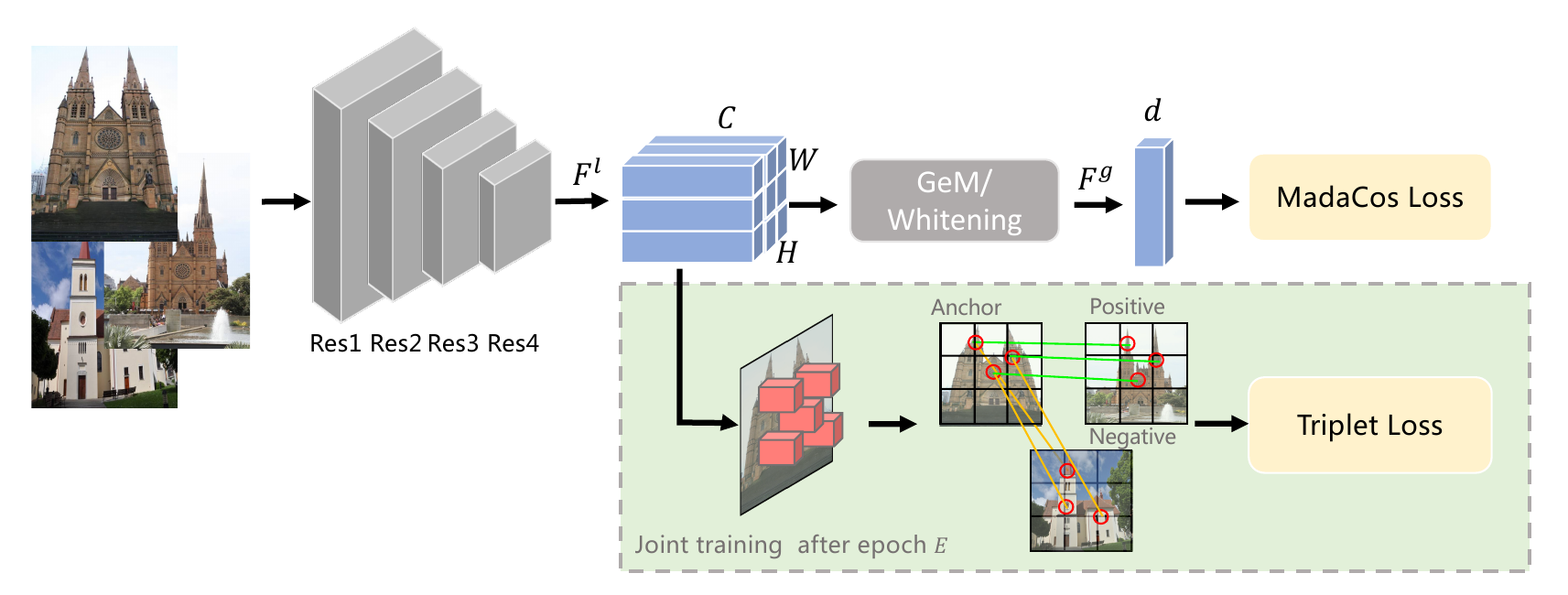}
  \caption{
    Illustration of the proposed coarse-to-fine framework to learn compact discriminative representation (CFCD) and its training objectives. The components highlighted in green are introduced after training for $E$ epochs. We use MadaCos alone to train global features for the first $E$ epochs, then select the attention regions from the attention maps as constraints to construct triplets for local descriptors matching, and finally train the model with both MadaCos and triplet losses. 
  }
  \label{fig:framework}
\end{figure*}

\subsection{Margin-based Softmax Loss}
For learning deep image representations, previous works propose pair-based losses such as  triplet~\cite{bronstein2010data}, angular~\cite{wang2017deep} and listwise~\cite{revaud2019learning} losses to train CNN models.
Recent works propose margin-based losses such as SphereFace\cite{liu2017sphereface}, CosFace\cite{wang2018cosface}, ArcFace\cite{deng2019arcface},and Sub-center ArcFace\cite{deng2020sub}, which maximize angular margin to the target logit and thus lead to faster convergence and better performance.
Among them, the state-of-the-art studies\cite{weyand2020google, yang2021dolg} directly adopt ArcFace loss to train the whole model.
However, the training process of ArcFace loss is usually tricky and unstable, so one has to repeat training with multiple settings to achieve optimal performance. Adacos\cite{zhang2019adacos} attempt to leverage adaptive or learnable scale and margin parameters, but they pay less attention to the softmax function curve and data characteristics, \eg large variations in background, occlusion and viewpoint. Therefore, we propose a MadaCos loss which automatically tunes its hyperparameters to perform more accurate single-stage image retrieval. 
\section{Methods}
\subsection{Overview}
Our CFCD framework is depicted in Fig.\ref{fig:framework}. Given an image, we obtain the original deep local descriptors $F^l \in \mathbb{R}^{d_c\times d_w\times d_h}$ via a CNN backbone, where $d_c$, $d_w$ and $d_h$ are the dimensions of channels, width and height of the feature map, respectively. We then use GeM pooling and a whitening FC layer to extract the global representation $F^g \in \mathbb{R}^{d_g}$ of a dimension $d_g$. We propose MadaCos, an adaptive softmax-based loss to learn the global representation. During the first $E$ training epochs, MadaCos is used alone to make the network aware of prominent local regions in $F^l$. The prominent regions are selected from the attention maps as matching constraints, which prompt us to design a hard negative sampling strategy to construct triplets for local descriptor matching. After $E$ epochs, a triplet loss is combined with MadaCos to jointly train the network so that the global features are infused with geometry information about discriminative local regions.

\subsection{Global Feature Learning with MadaCos Loss}
Let $x_i$ be the $i$-th image of the current mini-batch with size $N$, and $y_i$ the corresponding label. Margin-based softmax losses~\cite{liu2017sphereface,wang2018cosface,deng2019arcface} apply $\ell_2$ normalization to the classifier weight and the embedding feature. They use three kinds of margin penalty, $i.e.$, multiplicative angular margin $m_1$, additive angular margin $m_2$, and additive cosine margin $m_3$, respectively. We denote $\theta_{y_i}$ as the angle between the target weight and the feature, and $cos\theta_{y_i}$ is its cosine logit. 
Then the margin-based softmax losses can be combined into a unified framework:
\begin{equation}
\small
\mathcal{L}=-\frac{1}{N}\sum_{i=1}^N log\frac{e^{s(cos(m_1\theta_{y_i}+m_2)-m_3)}}{e^{s(cos(m_1\theta_{y_i}+m_2)-m_3)} + B_i}, 
\label{equation1}
\end{equation}
where $s$ is a scale factor, and $B_i=\sum_{j=1,j\ne y_i}^n e^{s\, cos\theta_j}$ is the summation of the cosine logits of non-target classes.

\begin{figure}[t]
  \centering
  \setlength{\abovecaptionskip}{-0.5mm}
  \includegraphics[width=1.0\linewidth, height=0.4\linewidth]
                  {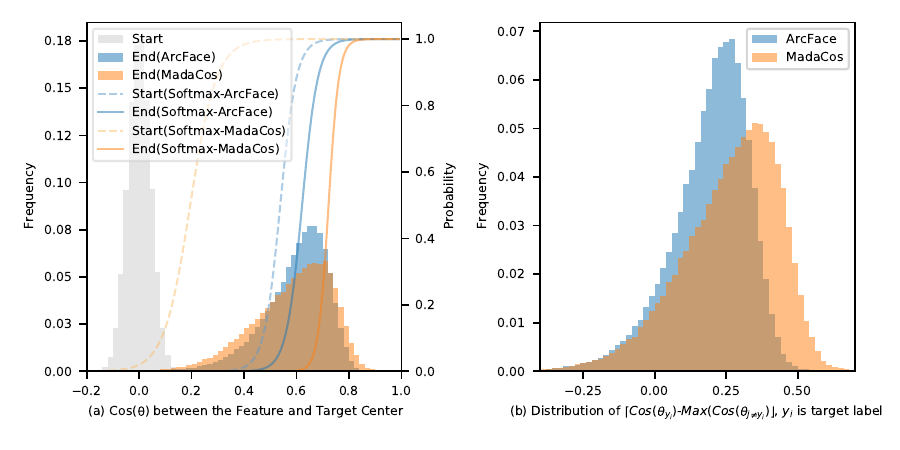}
  \caption{
    (a) Target cosine logit distributions and softmax function at the start and end of training. (b) Distributions of difference between the target cosine logit and its corresponding maximum of non-target cosine logit after training convergence.
  }
  \label{fig:distributions}
  \vspace{-1mm}
\end{figure}

Previous works\cite{weyand2020google, yang2021dolg} adopt ArcFace loss with additive angular margin $m_2$ to train the global descriptors, which achieves better performance than other margin-based softmax functions. 
According to Eq.\ref{equation1}, we show the distribution of $cos\theta_{y_i}$ between embedding feature and the corresponding target center as well as the softmax function curves at the start and end of training in Fig.\ref{fig:distributions}(a). As the training converges, $\theta_{y_i}$ gradually shrinks so its $cos\theta_{y_i}$ distribution and softmax function gradually shift to the right side.
ArcFace loss makes the distribution of class centers scattered and the distribution of $cos\theta_{y_i}$ more concentrated, which intuitively indicates that it enhances the intra-class compactness. 
Nevertheless, most samples distribute on the right side of the softmax function. This leads to suboptimal performance due to weak supervisions. We therefore propose a novel adaptive loss, namely MadaCos, which automatically tunes appropriate parameters within each mini-batch by imposing strict constrains on the probability of the median of cosine logits to progressively strengthen supervision throughout the training process.

To simplify the derivation of subsequent formulas, we adopt the additive cosine margin $m_3$ instead of additive angular margin $m_2$ to impose penalties, and the subscript in $m_3$ is omitted in the following for simplicity.
Formally, we have:
\begin{equation}
\small
P_{i,y_i}=\frac{e^{s(cos\theta_{y_i}-m)}}{e^{s(cos\theta_{y_i}-m)} + B_i}, 
\label{equation2}
\end{equation}
where $P_{i,y_i}$ represents its probability of assigning $x_i$ to class $y_i$. 
And we also introduce a modulating indicator variable $cos_m\theta$, which is the median of target cosine logits in the current mini-batch, and its corresponding label is $y_k$.
$cos_m\theta$ reflects the convergence degree of the under-training network in the the mini-batch.
Since the cross entropy loss is $-log(P)$, if we can control the probability $P(cos_m\theta)$ of the median target cosine logit, we potentially control the overall supervision intensity.
Therefore, we propose to dynamically compute the appropriate scale $s$ and margin $m$ within each mini-batch so that $P(cos_m\theta)$ reaches an anchor point $\rho$, which keeps most samples distributing on the left side of the softmax function. 
Accordingly, the network can impose stronger constraints to progressively enhance supervision with suitable $\rho$ even if the angle $\theta_{i,y_i}$ shrinks. Based on this observation, we set the $P(cos_m\theta)=\rho$ to compute scale $s$ and margin $m$. Here $\rho$ is set to 0.02 in our experiments,
\begin{equation}
\small
\frac{e^{s(cos_m\theta-m)}}{e^{s(cos_m\theta-m)} + \tilde B}=\rho, 
\label{equation3}
\end{equation}

Eq.\ref{equation3} ensures that the margin $m$ and scale $s$ gradually increase to avoid model divergence due to too large margin and scale at the early phase of training. 
However, if $s$ is too small ($e.g.$, $s = 10$), $P_{i,y_i}$ will be very low when $\theta_{y_i}$ is close to 0, which means that the loss function may still penalize correctly classified samples. Therefore, we force $P_{i,y_i}$ to be close to 1 when the angle $\theta_{y_i}$ is 0.
\begin{equation}
\small
\frac{e^{s(1 - m)}}{e^{s(1 - m)} + \tilde B}=1-\epsilon, 
\label{equation4}
\end{equation}
where $\tilde B = \sum_{j=1,j\ne y_k}^n e^{s\, cos\theta_j}$. We set $\epsilon=e^{-7}$ in our experiments. Combining Eq.\ref{equation3} and Eq.\ref{equation4}, we can derive $s$ and $m$ within each mini-batch to update the MadaCos loss to train the global descriptors progressively. The entire process is summarized in Algorithm \ref{algorithm1}. 

As shown in Fig.\ref{fig:distributions}(a), compared with ArcFace, the $cos\theta_{y_i}$ distributions of MadaCos more scattered and most samples are distributed on the left side of the softmax curve. We also plot the distribution of $[cos\theta_{y_i}-argmax(cos\theta_{j\ne y_i})]$ in Fig.\ref{fig:distributions}(b), which intuitively illustrates that MadaCos loss has a larger tolerance margin. 

\begin{algorithm}[tb]
\small
\caption{MadaCos}
\label{algorithm1}
\textbf{Input}: The image $x_i$ of $i$-th sample with label $y_i$ in the mini-batch with size $N$, the target cosine logit $cos\theta_{y_i}$ \\ 
\textbf{Parameter}: scale $s$ and margin $m$
\begin{algorithmic}[1] 
\STATE $cos_m\theta=Median\{cos\theta_{y_0},cos\theta_{y_1},\ldots,cos\theta_{y_{N-1}}\}$;
\STATE Substitute $cos_m\theta$ into Eq.\ref{equation3} and Eq.\ref{equation4} to compute scale $s$; 
$$s = \frac{log((1-\epsilon)(1-\rho)/(\rho\epsilon))}{1-cos_m\theta}$$
\STATE Assuming that the corresponding label of $cos_m\theta$ is $y_k$, substitute $s$ to compute $\tilde B_k=\sum_{j\ne y_k}^n e^{s\, cos\theta_j}$;
\STATE Compute the margin $m$ with scale $s$ and $\tilde B_k$ by Eq.\ref{equation4};
$$m = cos_m\theta - \frac{log(\rho\tilde B_k /(1-\rho))}{s}$$
\STATE Update the loss $\mathcal{L}_{mda}$ with $m$ and $s$;
$$\mathcal{L}_{mda}=-\frac{1}{N}\sum_{i=1}^N log\frac{e^{s(cos\theta_{y_i}-m)}}{e^{s(cos\theta_{y_i}-m)} + B_i}$$
\end{algorithmic}
\textbf{Output}: loss $\mathcal{L}_{mda}$.
\end{algorithm}

\subsection{Local Feature Matching with Triplet Loss}

Let $f \in \mathbb{R}^{d_c}$ be a local descriptor from local descriptors $F^l$, and then $F^l$ can be seen as set of $Z=d_w \times d_h$ feature vectors denoted by $\mathcal{F}=\{f_i \in \mathbb{R}^{d_c}: i \in 1 \ldots Z\}$. We select prominent local descriptors to minimize their pairwise distance. Obviously, if we select matching local descriptors only relying on nearest neighbor-based constraints, the local descriptors may focus on the non-significant parts such as backgrounds and distractions. 
Therefore, we introduce attention maps to guide the networks to focus on more semantically salient regions and discard the redundant information. Here we define the function $\eta(f, \mathcal{F})=argmin_{\forall f_i\in \mathcal{F}}||f-f_i||_2$ which returns the nearest neighbor of $f$ from set $\mathcal{F}$. 
And let $\psi(\tau,\mathcal{F})=\{argmax^{\tau}_{\forall f_i\in \mathcal{F}} ||f_i||\}$ be the attention selective function that returns the top $\tau$ percent local descriptors by $l_1$ norm, where $\tau$ is a controlling factor. 
Now, given a positive pair of images $x_a, x_p$, the corresponding local descriptors are $F_a^l$ and $F_p^l$. For any eligible local descriptors $v\in \mathcal{F}$, in order to select matching descriptors $(v_a, v_p)$ of positive pair, we set the following constraints: a) $(v_a, v_p)$ must be reciprocal nearest neighbors, b) they need to be in the attention regions specified by $\psi(\tau,\mathcal{F})$. Let $\mathcal{M}$ be the set of eligible pairs, the conditions above can be formulated as:
\begin{equation}
\small
(v_a, v_p)\in \mathcal{M} \Longleftrightarrow 
\begin{cases} 
v_a = \eta(v_p, \psi(\tau, F_a^l))  \\
v_p = \eta(v_a, \psi(\tau, F_p^l))
\end{cases} 
\label{equation5}
\end{equation}

Once we construct all eligible local descriptors $\mathcal{M}$ of positive pairs, we introduce the triplet loss to leverage rich local relations between matches.
For $Q$ negative images $x_{n_j}, j=1,\ldots,Q$, we only rely on the nearest neighbor criterion to select those closest to the anchor image matches. 
We denote $v_{n_j}=\eta(v_a, F_{n_j}^l)$ as the negative local descriptors extracted from $x_{n_j}$. The triplet loss can be written as:
\begin{equation}
\small
\mathcal{L}_{trip}=\sum_{\tiny{(v_a,v_p)\in \mathcal{M}}}\sum_{j=1}^Q\{||v_a-v_p||_2^2-||v_a-v_{n_j}||_2^2+\mu\}^+, 
\label{equation6}
\end{equation}
where $\mu=0.1$ and $Q=6$ in our experiments.
 
However, triplets constructed by random sampling do not provide sufficiently strong supervision for this task. 
To not only keep the accurate prediction of the normal and easy samples, but also make the model concentrate on learning from hard samples, we design a custom global sampling strategy with hard negative samples. 
The approach is shown in Fig.\ref{fig:sampling}, and the detailed sampling strategy is provided in the supplementary material. 
With the help of the model trained at a sufficient stage, we can use its prediction to ensure that each batch of triplets contain appropriate positives which share common patch-level matches between them while focusing on hard negatives. 
The sampling strategy is crucial to select hard negative triplets and therefore contributes to improving the overall performance. Unlike previous work \cite{lee2022cvnet} which selects negatives in the order of global descriptor matching scores with MoCo-like \cite{he2020momentum} momentum queue, we select negatives from the whole dataset at each epoch without additional computation.

Finally, the total loss of our backbone network $\mathcal{L}_{tot}$ is the weighted sum of the classification loss $\mathcal{L}_{mda}$ and triplet loss $\mathcal{L}_{trip}$:
\begin{equation}
\small
\mathcal{L}_{tot}=\mathcal{L}_{mda} + \lambda \mathcal{L}_{trip}, 
\label{equation7}
\end{equation}
where $\lambda$ is set to 0.05 during training.

\begin{figure}[t]
  \centering
  \setlength{\abovecaptionskip}{-0.5mm}
  \includegraphics[width=1.0\linewidth, height=0.5\linewidth]
                  {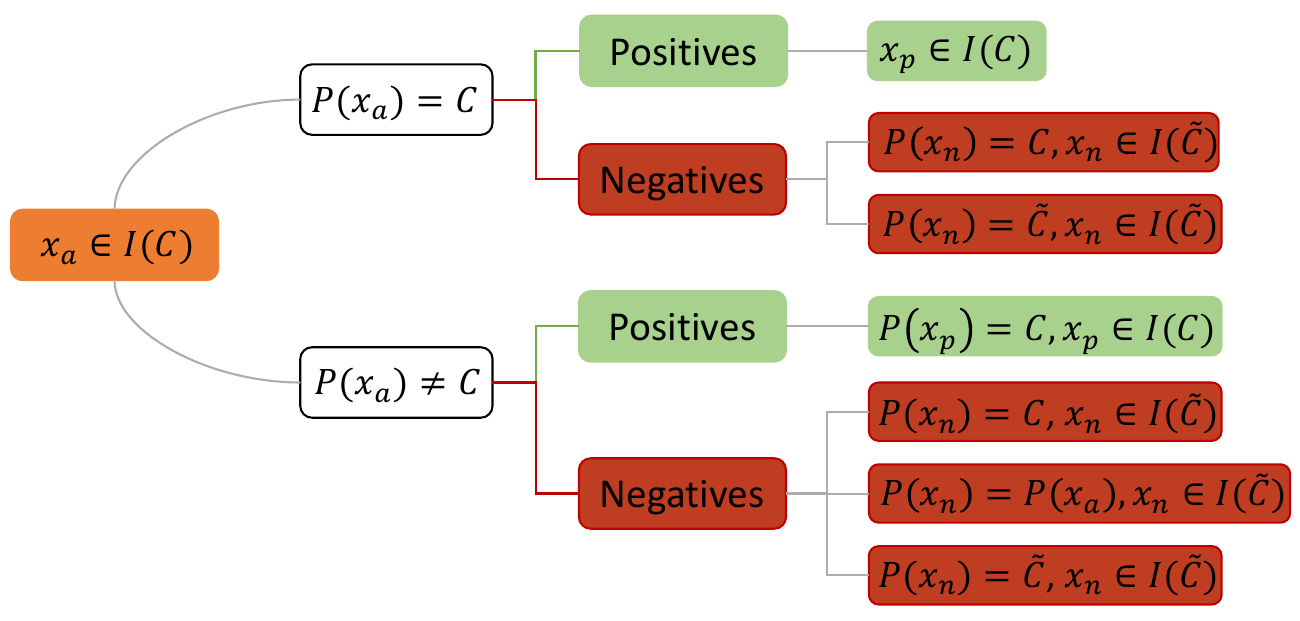}
  \caption{
    Hard negative sampling strategy. Given an anchor image $x_a$ with category $C$ and its prediction $P(x_a)$. $I(C)$ and $I(\tilde{C})$ are the sets of images with category $C$ and non-category $\tilde{C}$, respectively. If $P(x_a)=C$, we randomly sample positives from set $I(C)$ and evenly select negatives from the two predictions of images $x_n\in I(\tilde{C})$. However, if $P(x_a)\neq C$, the anchor image itself is a hard or noise sample, we require that the positives must be select from set $I(C)$ and it's prediction $P(x_p)=C$. The negatives are also evenly selected from the three predictions of images $x_n\in I(\tilde{C})$.
  } 
  \label{fig:sampling}
  \vspace{-2mm}
\end{figure}

\begin{table*}
    \small
	\centering
	\renewcommand\arraystretch{1.2}
	\setlength{\tabcolsep}{2.8pt}
	\begin{tabular}{lccccccccccccc}
		\hline
		\multirow{2}{*}{Method} & \multicolumn{4}{c}{Medium} &
		\multicolumn{4}{c}{Hard} &  & \multicolumn{2}{c}{Multi-scale} \\
		\cline{2-9} 
            \cline{11-12}
		 & $\mathcal{R}$Oxf & +1M  & $\mathcal{R}$Par  & +1M & 
		$\mathcal{R}$Oxf & +1M  & $\mathcal{R}$Par &  +1M &  & scale & dimen\\
		\hline
		(A) $Local\ features\ aggregation\ +\ re$-$ranking$   \\
		\hdashline
		HesAff-rSIFT-ASMK$^*$+SP\cite{tolias2016image}  & 60.60  & 46.80  & 61.40 & 42.30 & 36.70 & 26.90 & 35.00 & 16.80 & & - & -\\
		DELF-ASMK$^*$+SP(GLDv1)\cite{noh2017large, radenovic2018revisiting}  & 67.80  & 53.80  & 76.90 & 57.30 & 43.10 & 31.20 & 55.40 & 26.40 & & - & -\\
		DELF-D2R-R-ASMK$^*$+SP(GLDv1)\cite{teichmann2019detect}  & 76.00  & 64.00  & 80.20 & 59.70 & 52.40 & 38.10 & 58.60 & 29.40 & & - & -\\
		R50-How-ASMK,n=2000\cite{tolias2020learning}   & 79.40 & 65.80 & 81.60 & 61.80 & 56.90 & 38.90 & 62.40 & 33.70 & & - & -\\
            FIRe(SfM-120k)\cite{superfeatures}   & $\mathbf{81.80}$  & $\mathbf{66.50}$  & $\mathbf{85.30}$ & $\mathbf{67.60}$ & $\mathbf{61.20}$ & $\mathbf{40.10}$ & $\mathbf{70.00}$ & $\mathbf{42.90}$ & & 7 & -\\
		\hline

		(B) $Global\ features\ +\ Local\ feature\ re$-$ranking$   \\
		\hdashline
		R101-GeM+DSM \cite{simeoni2019local}  & 65.30  & 47.60 & 77.40 & 52.80 & 39.20 & 23.20 & 56.20 & 25.00 & & - & -\\
		R50-DELG(GLDv2-clean)\cite{cao2020unifying}  & 78.30  & 67.20 & 85.70 & 69.60 & 57.90 & 43.60 & 71.00 & 45.70 & & 3 & 2048\\
		R101-DELG(GLDv2-clean)\cite{cao2020unifying}  & 81.20  & 69.10  & 87.20 & 71.50 & 64.00 & 47.50 & 72.80 & 48.70 & & 3 & 2048\\
  		R50-CVNet-Rerank(Top-400)\cite{lee2022cvnet} & $\mathbf{ 87.90}$  &  80.70 & 90.50 & 82.40 & 75.60 & 65.10 & 80.20 & 67.30 & & 3 & 2048\\
  		R101-CVNet-Rerank(Top-400)\cite{lee2022cvnet} & 87.20  & $\mathbf{ 81.90}$ & $\mathbf{91.20}$ & $\mathbf{ 83.80}$ & $\mathbf{ 75.90}$ & $\mathbf{ 67.40}$ & $\mathbf{81.10}$ & $\mathbf{ 69.30}$ & & 3 & 2048\\
		\hline
  		(C) $Global\ features$   \\
		\hdashline
		R101-SOLAR(GLDv1)\cite{ng2020solar}  & 69.90   & 53.50  & 81.60 & 59.20 & 47.90 & 29.90 & 64.50 & 33.40 & & 3 & 2048\\
  		R50-DOLG(GLDv2-clean)$^r$\cite{yang2021dolg}  & 80.05   & 70.53  & 89.49 & 77.85 & 60.75 & 44.63 & 77.45 & 57.52 & & 5 & 512\\
		R101-DOLG(GLDv2-clean)$^r$\cite{yang2021dolg}  & $\mathbf{81.97}$   & 72.43  & 90.11 & 80.24 & $\mathbf{63.76}$ & 48.28 & 78.20 & 61.33 & & 5 & 512\\
  		R50-CVNet-Global(GLDv2-clean)\cite{lee2022cvnet} & 81.00 & 72.60 & 88.80 & 79.00 & 62.10 & 50.20 & 76.50 & 60.20 & & 3 & 2048\\
		R101-CVNet-Global(GLDv2-clean)\cite{lee2022cvnet} & 80.20 & $\mathbf{74.00}$ & $\mathbf{90.30}$ & $\mathbf{80.60}$ & 63.10 & $\mathbf{53.70}$ & $\mathbf{79.10}$ & $\mathbf{62.20}$ & & 3 & 2048\\
		\hline
            R50-CFCD(GLDv2-clean)  & $\mathbf{82.51}$   & $\mathbf{72.73}$  & $\mathbf{89.64}$ & $\mathbf{78.91}$ & $\mathbf{63.59}$ & $\mathbf{48.54}$ & $\mathbf{78.06}$ & $\mathbf{60.09}$ & & 3 & 2048\\
		R101-CFCD(GLDv2-clean)  & $\mathbf{84.08}$   & $\mathbf{74.66}$  & $\mathbf{91.03}$ & $\mathbf{82.18}$ & $\mathbf{67.80}$ & $\mathbf{54.10}$ & $\mathbf{81.21}$ & $\mathbf{65.51}$ & & 3 & 2048\\
		R50-CFCD(GLDv2-clean)  & $\mathbf{82.42}$   & $\mathbf{73.06}$  & $\mathbf{91.57}$ & $\mathbf{81.57}$ & $\mathbf{65.06}$ & $\mathbf{50.78}$ & $\mathbf{81.69}$ & $\mathbf{62.80}$ & & 5 & 512\\
		R101-CFCD(GLDv2-clean)  & $\mathbf{85.24}$   & $\mathbf{73.99}$  & $\mathbf{91.56}$ & $\mathbf{82.80}$ & $\mathbf{69.96}$ & $\mathbf{52.78}$ & $\mathbf{81.78}$ & $\mathbf{65.78}$ & & 5 & 512\\
		\hline
	\end{tabular}
	\caption{Results (\% mAP) of different methods on $\mathcal{R}$Oxf(+1M) and $\mathcal{R}$Par(+1M) with Medium and Hard evaluation protocols. State-of-the-art performances are marked bold and our results are summarized in the bottom section. ``$*$'' means feature quantization is used. Methods superscripted with $^r$ are our re-implementations. ``scale'' and ``dimen'' are different scales and dimensions for global features. Our method belongs to the global features single pass group (C).} 

	\label{table1}
	\vspace{-1mm}
\end{table*}

\section{Experiments}
\subsection{Implementation Details}

{\bf Datasets and Evaluation Metric} The clean version of Google landmarks dataset V2 (GLDv2-clean) ~\cite{weyand2020google} contains 1,580,470 images and 81,313 classes. We randomly divide $80\%$ of the data for training and the rest $20\%$ for validation following previous works\cite{cao2020unifying, yang2021dolg}. To evaluate our model, we primarily use $\mathcal{R}$Oxford5k \cite{philbin2007object, radenovic2018revisiting} and $\mathcal{R}$Paris6k \cite{philbin2008lost, radenovic2018revisiting} datasets, denoted as $\mathcal{R}$Oxf and $\mathcal{R}$Par. Both datasets comprise 70 queries and include 4993 and 6322 database images, respectively. In addition, an $\mathcal{R}$1M dataset \cite{radenovic2018revisiting} which contains one million distractor images is used for measuring the large-scale retrieval performance. For a fair comparison, mean average precision (mAP) is used as our evaluation metric on both datasets with the medium and hard difficulty protocols. 

{\bf Training Details} ResNet50 and ResNet101\cite{he2016deep} are mainly used for experiments. Models in this paper are initialized from Imagenet\cite{deng2009imagenet} pre-trained weights. The images first undergo augmentations include random cropping and aspect ratio distortion, then are resized to 512$\times$512 following previous works\cite{cao2020unifying, yang2021dolg}. The models are trained on 8 V100 GPUs for $T$ epochs with the batch size of 128. The initial learning rate of is 0.01. We use SGD optimizer with momentum of 0.9, and set weight decay factor to 0.0001. We also adopt the cosine learning rate decay strategy in the first $E$ epochs to train with MadaCos, and after the $E$-th epoch, we reset the learning rate to 0.005 to continue training the model with both MadaCos and triplet losses for the remaining $T-E$ epochs. For GeM pooling, we fix the parameter $\boldsymbol{p}$ as 3.0. 
As for global feature extraction, we also produce multi-scale representations. And $\ell_2$ normalization is applied for each scale independently then they are average-pooled, followed by another $\ell_2$ normalization step.
We use two kinds of experimental settings for fair comparisons. For comparing with DOLG and FIRe\cite{superfeatures}, we set $T$ to 100, $E$ to 50, $\boldsymbol{d_g}$ to 512 and use 5 scales, $\{\frac{1}{2\sqrt{2}}, \frac{1}{2}, \frac{1}{\sqrt{2}}, 1, \sqrt{2}\}$. 
For comparing with other methods, we set $T$ to 25, $E$ to 20, $\boldsymbol{d_g}$ to 2048 and use 3 scales, $\{\frac{1}{\sqrt{2}}, 1, \sqrt{2}\}$.

\subsection{Results}
In Tab.\ref{table1}, we divide the previous methods into three groups: (A) local features aggregation and re-ranking; (B) global features followed by local features re-ranking; and (C) global features. Our CFCD belongs to the group C. It can be observed that our solution consistently outperforms existing one-stage methods without additional computation overhead. 

{\bf Comparison with One-stage State-of-the-art Methods.}
{\bf 1)} Like methods in the global feature based group C, our method performs single-stage image retrieval with only the global feature.
Due to the misreported results\footnote{\href{https://github.com/feymanpriv/DOLG-paddle/issues/3}{https://github.com/feymanpriv/DOLG-paddle}} in DOLG , we re-implement the R50/101-DOLG$^r$ in the official configuration and achieve similar performance.
R50/101-DOLG$^r$ using local branch and orthogonal fusion module to combine both local and global information is still an excellent single-stage method.
Our R50-CFCD with ResNet50 backbone even outperform R101-DOLG$^r$ with ResNet101 backbone in all settings. Notably, our method R101-CFCD outperforms the R101-DOLG$^r$ with a gain of up to 3.27$\%$ on $\mathcal{R}$Oxf-Medium, 1.45$\%$ on $\mathcal{R}$Par-Medium, 6.2$\%$ on $\mathcal{R}$Oxf-Hard and 3.58$\%$ on $\mathcal{R}$Par-Hard.
Even with a large amount of distractors in the database, our R50-CFCD and R101-CFCD still outperform R50-DOLG$^r$ and R101-DOLG$^r$ by a large margin, respectively, which demonstrates the effectiveness of our method to exploit fine-grained local information.
{\bf 2)} When compared with CVNet-Global, the proposed CFCD exhibits significantly superior performance in almost all dataset.
These results exhibit excellent performance of our framework on single-stage image retrieval benchmarks. It should be noted that our method does not contain the expensive local attention module as in \cite{yang2021dolg}. This suggests that infusing aligned local information into the final descriptor is a better option.

{\bf Comparison with Other Two-stage Methods.}
{\bf 1)} In the local feature based group A, FIRe \cite{superfeatures} is the current state-of-art local feature aggregation method and it outperforms R50-How-ASMK. Regardless of its complexity, our R50-CFCD outperforms it by 3.86$\%$ on Roxf-Hard and 11.69$\%$ on Rpar-Hard with the same ResNet50 backbone. 
{\bf 2)} In group B, another type of two-stage methods are based on the retrieve-and-rerank paradigm where the global retrieval is followed by a local feature re-ranking. Such methods exhibit superior performance owing to the nature of re-ranking. But our one-stage method (R50-CFCD) still outperforms the two-stage method (R101-DELG) on the both $\mathcal{R}$Oxf dataset and $\mathcal{R}$Par dataset by a significant margin.
However, the re-ranking network of CVNet is trained with 1M images manually selected from 31k landmarks of the GLDv2-clean dataset. 
Since the authors of CVNet\cite{lee2022cvnet} do not upload their cleaned training data, it should be noted that comparisons in Tab.\ref{table1} is not fair for our method.
More comparisons with other two-stage methods are provided in the supplementary material.

\begin{figure*}[t]
  \centering
  \setlength{\abovecaptionskip}{-0.5mm}
  \includegraphics[width=1.0\linewidth, height=0.46\linewidth]
                  {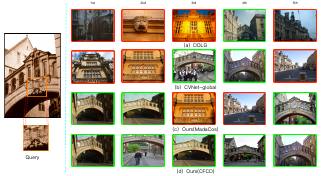}
  \caption{
    Demonstration of the top-5 retrieved results. The query on the left used as an input is generated by cropping only the part bounded by a orange box. On the right are the results of DOLG, CVNet-global and Ours(only Madacos loss and CFCD methods), which are shown from top to bottom. Green and red boxes denote positive and negative images, respectively. 
  }
  \label{fig:samples}
  \vspace{-2mm}
\end{figure*}

{\bf Qualitative Analysis.} We showcase top-5 retrieval results with a hard query image in Fig.\ref{fig:samples} of different methods. We can observe many false positives in the retrieval list of DOLG and CVNet-global, because its weak and implicit supervision on the local information is not robust enough when the query information is focused on a local patch. In contrast, more true positives can be recalled with only global features trained by our MadaCos loss. When additionally introducing triplet loss with the matching strategy to integrate local information into the global features, we obtain more robust retrieval results.

\subsection{Ablation Studies}
In this section, we conduct ablation experiments using the ResNet50 backbone to empirically validate the components of CFCD. We use 5 scales, $\{\frac{1}{2\sqrt{2}}, \frac{1}{2}, \frac{1}{\sqrt{2}}, 1, \sqrt{2}\}$ and set the dimension $\boldsymbol{d_g}$ of the global feature to 512.

\begin{table}
    \small
	\centering
	\renewcommand\arraystretch{1.2}
	\setlength{\tabcolsep}{2.3pt}
	\begin{tabular}{ccccccccc}
		\hline
		\multirow{2}{*}{$\#$} & \multirow{2}{*}{MadaCos} & \multirow{2}{*}{Triplet} & \multirow{2}{*}{HNS} &  \multirow{2}{*}{$E$} & \multicolumn{2}{c}{Medium} &
		\multicolumn{2}{c}{Hard} \\
		\cline{6-9}
		& & & & & $\mathcal{R}$Oxf & $\mathcal{R}$Par & 
		$\mathcal{R}$Oxf & $\mathcal{R}$Par\\
		\hline
		 1 & & & & - & 78.76 & 88.59 & 58.53 & 75.97 \\
		 2 & $\checkmark$ & & & - & 81.86 & 90.80 & 64.38 & 80.21 \\
		 3 & $\checkmark$ & $\checkmark$ & & 50 & 82.04 & 91.23 & 64.84 & 81.68 \\
		 4 & $\checkmark$ & $\checkmark$ & & 0 & 78.18 & 89.58 & 58.29 & 75.79 \\
		 5 &  & $\checkmark$ & $\checkmark$ & 50 & 79.96 & 89.54 & 60.46 & 77.61 \\
		 6 & $\checkmark$ & $\checkmark$ & $\checkmark$ & 50 & $\mathbf{82.42}$ & $\mathbf{91.57}$ & $\mathbf{65.06}$ & $\mathbf{81.69}$ \\
		\hline
	\end{tabular}
	\caption{Ablation study on different components. ``MadaCos'' means our median adaptive loss. ``Triplet'' means training with the triplet loss. ``HNS'' means the hard negative sampling strategy. ``$E$'' is the epoch when triplet loss is added for training.} 

	\label{table2}
	\vspace{-2mm}
\end{table}

\begin{table}
    \small
	\centering
	\renewcommand\arraystretch{1.2}
	\begin{tabular}{cccccc}
		\hline
		\multirow{2}{*}{Loss} &
		\multirow{2}{*}{$\rho$} &
		\multicolumn{2}{c}{Medium} &
		\multicolumn{2}{c}{Hard} \\
		\cline{3-6}
		& & $\mathcal{R}$Oxf & $\mathcal{R}$Par & 
		$\mathcal{R}$Oxf & $\mathcal{R}$Par\\
		\hline
		ArcFace~\cite{deng2019arcface} & - & 78.76 & 88.59 & 58.53 & 75.97 \\
    	AdaCos\cite{zhang2019adacos} & - & 75.88 & 87.44 & 56.62 & 73.99 \\
		\hline
		\multirow{5}{*}{MadaCos} 
		 & 0.01 & 80.44 & 89.46 & 62.36 & 77.57 \\
		 & 0.02 & $\mathbf{81.78}$ & $\mathbf{90.60}$ & 63.36 & $\mathbf{79.94}$ \\
		 & 0.03 & 80.87 & 89.62 & 62.30 & 78.46 \\
		 & 0.04 & 81.63 & 89.33 & $\mathbf{64.90}$ & 77.13 \\
		 & 0.05 & 81.72 & 89.51 & 63.18 & 77.51 \\
		\hline
		CosFace~\cite{wang2018cosface} & 0.02 & 71.01 & 82.39 & 50.28 & 68.7 \\
		\hline
	\end{tabular}
	\caption{Ablation study on different $\rho$ in MadaCos function.}
	\label{table3}
	\vspace{-2mm}
\end{table}

\begin{figure}[t]
  \centering
  \setlength{\abovecaptionskip}{-0.5mm}
  \includegraphics[width=1.0\linewidth]
                  {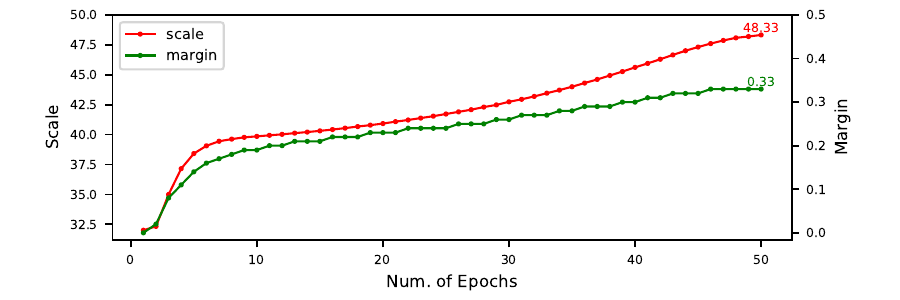}
  \caption{
    The values of the scale and margin of MadaCos during training with $\rho=0.02$. 
  }
  \label{fig:scales_margins}
  \vspace{-1mm}
\end{figure}

{\bf Verification of Different Components.} In Tab.\ref{table2}, we provide detailed ablation experimental results, verifying the contributions of three components in the framework by incrementally adding components to the baseline framework. 
For the baseline framework in the first row, we set the ArcFace margin $m$ as 0.15 and scale $s$ as 30 following DOLG, and train the model with ArcFace loss for 100 epoch. 
When the MadaCos loss is adopted to train for 100 epoch, mAP increases from 78.76$\%$ to 81.86$\%$ on $\mathcal{R}$Oxf-Medium and from 58.53$\%$ to 64.38$\%$ on $\mathcal{R}$Oxf-Hard. 
Then we introduce the coarse-to-fine framework to train the whole model with MadaCos and triplet losses, and observe that selecting local descriptors to discover patch-level matches between images helps to improve the overall performance, especially on hard cases. The mAP is improved from 64.38$\%$ to 64.84$\%$ on $\mathcal{R}$Oxf-Hard and from 80.21$\%$ to 81.68$\%$ on $\mathcal{R}$Par-Hard as in row 3. This indicates that aligning matching local descriptors according to visual patterns makes them more discriminative.
The comparisons between 1st and 5th rows shows that the performance of ArcFace can be significantly improved by the coarse-to-fine framework.
However, the comparisons between 3rd and 4th rows suggest that naively optimizing with total loss from the start leads to suboptimal performance, because during early training stage the local features are too premature for feature matching and may damage the global feature representation. 
In the last row, the performance is further improved when training with hard samples.

{\bf Loss Comparison for Global Feature Learning.} The results of training with different anchor point $\rho$ in softmax function are shown in Tab.\ref{table3}. Unlike the scale $s$ and margin $m$ of the ArcFace, the single anchor point $\rho$ is the only manually tunable parameter in MadaCos. We simply adjust $\rho$ from 0.01 to 0.05 to train the global descriptors for only 50 epochs, and the results are significantly improved, even surpassing R50-DOLG$^r$ which is trained for 100 epochs.

As $\rho$ increases, the mAP performance first increases and then decreases, with the best results at $\rho=0.02$. Fig.\ref{fig:scales_margins} illustrates the value change of scale $s$ and margin $m$ in MadaCos. As the training proceeds, the scale $s$ and margin $m$ gradually increase and then plateau out. In the last row of Tab.\ref{table3}, as we fix $s=48.33, m=0.33$ according to their converged values at the 50th epoch in Fig.\ref{fig:scales_margins}, MadaCos then degenerates into the initial CosFace, where the model performance drops sharply. 
This indicates that training with a large scale and margin at the beginning provides suboptimal supervision. Compared to existing parameter-free loss as AdaCos\cite{zhang2019adacos}, our dynamically tuned scaling parameters are more efficient. More experiments on the face recognition datasets are provided in the supplementary material.

\begin{table}
    \small
	\centering
	\renewcommand\arraystretch{1.2}
	\setlength{\tabcolsep}{5pt}
	\begin{tabular}{cccccc}
		\hline
		\multirow{2}{*}{Config} & \multirow{2}{*}{Layer} & \multicolumn{2}{c}{Medium} &
		\multicolumn{2}{c}{Hard} \\
		\cline{3-6}
		 & & $\mathcal{R}$Oxf & $\mathcal{R}$Par & 
		$\mathcal{R}$Oxf & $\mathcal{R}$Par\\
		\hline
  		w/o matching & $Res4$ & 80.97 & 91.22 & 62.37 & 80.51 \\
		matching & $Res4$ & $\mathbf{82.42}$ & $\mathbf{91.57}$ & $\mathbf{65.06}$ & $\mathbf{81.69}$ \\
		\hline
	\end{tabular}
	\caption{Experiments of triplet loss with matching constraints. ``w/o matching'' means introducing triplet loss without matching constraints to the global descriptors before the whitening FC layer, ``$Res4$'' means selecting local descriptors based on Res4.}
	\label{table4}
	\vspace{-1mm}
\end{table}

{\bf Impact of Triplet Loss with Matching Constraints.} 
We also provide experimental results to validate the impact of triplet loss with matching constraints. Our coarse-to-fine framework adopts MadaCos and triplet losses with different configurations to train model for 100 epochs, and the results are summarized in Tab.\ref{table4} and Tab.\ref{table5}.  In Tab.\ref{table4}, we can observe introducing the triplet loss with matching strategy to learn semantically salient region relations at $Res4$ improves the overall performance. We further explore the impact of controlling factor $\tau$ at $Res4$ in Tab.\ref{table5} which indicates smaller $\tau$ increases the overall performance. This is because selecting matching background information may bring noise to the global features, and stricter matching constraints can help the global features integrate more discriminative local information.

\begin{table}
    \small
	\centering
	\renewcommand\arraystretch{1.2}
	\setlength{\tabcolsep}{10pt}
	\begin{tabular}{ccccc}
		\hline
		\multirow{2}{*}{$\tau$(\%)} & \multicolumn{2}{c}{Medium} &
		\multicolumn{2}{c}{Hard} \\
		\cline{2-5}
		 & $\mathcal{R}$Oxf & $\mathcal{R}$Par & 
		$\mathcal{R}$Oxf & $\mathcal{R}$Par\\
		\hline
		30 & $\mathbf{82.42}$ & $\mathbf{91.57}$ & 65.06 & $\mathbf{81.69}$ \\
		50 & 82.02 & 91.41 & $\mathbf{65.10}$ & 81.46 \\
		70 & 81.88 & 91.35 & 64.94 & 81.38 \\
		\hline
	\end{tabular}
	\caption{Experimental results of triplet loss with different controlling factors $\tau$.}
	\label{table5}
	\vspace{-1mm}
\end{table}

\section{Conclusion}
In this paper, we propose $\mathbf{C}$oarse-to-$\mathbf{F}$ine framework to learn $\mathbf{C}$ompact and $\mathbf{D}$iscriminative representation (CFCD), an end-to-end image retrieval framework which dynamically tunes the hyperparameters of its loss function progressively to strengthen supervision for improving intra-class compactness and leverages fine-grained semantic relations to infuse global feature with inter-class distinctiveness. The resulting framework is robust to local region variations as well as exhibits more potential to real-world applications due to its single-stage inference without additional computation overhead of local feature re-ranking. Extensive experiments demonstrates the effectiveness and efficiency of our method, which provides a practical solution to difficult retrieval tasks such as landmark recognition.

{\small
\bibliographystyle{ieee_fullname}
\bibliography{egbib}
}

\newpage
\clearpage












\appendix

\noindent{\Large \textbf{Appendix}}
\section{More Implementation details}
\subsection{Details of Hard Negative Sampling Strategy}

For any epoch after the $E$-th epoch, we construct triplets of anchor, positive and negative samples for each class, and select them from the whole dataset based on the predictions of the trained model at the $E$-th epoch. Let $I(C)$ and $I(\tilde{C})$ be the sets of images with category $C$ and non-category  $\tilde{C}$ (\ie all categories other than $C$), respectively. 
We randomly sample an anchor image $x_a$ with category $C$ and get its prediction $P(x_a)$ as supervision.
If $P(x_a)=C$, we randomly sample one positive image $x_p$ from set $I(C)$, and evenly select $Q$ negative images from two types of negatives in set $I(\tilde{C})$, \ie, normal negatives with $P(x_n)=\tilde{C}$ and hard negatives with $P(x_n)=C$. Here, $x_n$ denotes the negative image. 
However, if $P(x_a)\neq C$, the anchor image itself is a hard or noise sample. Then we enforce that the positive image must be selected from $I(C)$ while satisfying $P(x_p)=C$ in order to avoid the situation where both $x_a$ and $x_p$ are noise samples. 
We also evenly select negatives from three types of negatives in set $I(\tilde{C})$, \ie, normal negatives with $P(x_n)=\{\tilde{C} - P(x_a)$\}, hard negatives with $P(x_n)=C$ and hard negatives with $P(x_n)=P(x_a)$, respectively. 
By learning with the well-constructed triplets, the network can further improve the overall performance.The entire process
is summarized in Algorithm \ref{algorithm2}.

\begin{algorithm}[tb]
\small
\caption{Hard Negative Sampling Strategy}
\label{algorithm2}
\textbf{Input}: The whole dataset $X$ with ground truth labels $G$ and the predictions $P$ of the trained model at the $E$-th epoch. \\
\textbf{Define}: $x\Leftarrow I$, randomly sample image $x$ from set $I$ without replacement; $rand()$ returns the random value uniformly sampled between $[0,1]$. \\
\textbf{Parameter}: The number of negatives $Q$, the collection of all classes $\mathcal{C}_n$ and the final training list $Y$. 
\begin{algorithmic}[1] 
\WHILE{$X \ne \varnothing$}
    \FOR {$C$ in $\mathcal{C}_n$}
        \STATE $I(C)=\{\forall x\in X\quad s.t.\quad G(x)=C\}$;
        \STATE $I(\tilde{C})=X-I(C)$;
        
        \STATE $x_a\Leftarrow I(C)$;
        \IF {$P(x_a)=C$}
            \STATE $x_p\Leftarrow I(C)$;
            \FOR {$j=1$ to $Q$}
                \IF {$rand() \leq$ 0.5}
                    \STATE $x_{n_j}\Leftarrow I(\tilde{C})\quad s.t.\quad P(x_{n_j})=C$; 
                \ELSE
                    \STATE $x_{n_j}\Leftarrow I(\tilde{C})\quad s.t.\quad P(x_{n_j})=\tilde{C}$; 
                \ENDIF
            \ENDFOR
        \ELSE
            \STATE $x_p\Leftarrow I(C)\quad s.t.\quad P(x_p)=C$;
            \FOR{$j=1$ to $Q$}
                \IF {$rand() \leq \frac{1}{3}$}
                    \STATE $x_{n_j}\Leftarrow I(\tilde{C})\quad s.t.\quad P(x_{n_j})=C$; 
                \ELSIF {$rand() \ge \frac{2}{3}$}
                    \STATE $x_{n_j}\Leftarrow I(\tilde{C})\quad s.t.\quad P(x_{n_j})=P(x_a)$;
                \ELSE
                    \STATE $x_{n_j}\Leftarrow I(\tilde{C})\quad s.t.\quad P(x_{n_j})=\{\tilde{C} - P(x_a)\}$; 
                \ENDIF
            \ENDFOR
        \ENDIF
    \ENDFOR
    \STATE $Y.$append($\{x_a, x_p, x_{n_1},\dots,x_{n_Q}\}$)
\ENDWHILE
\end{algorithmic}
\textbf{Output}: $Y$.
\vspace{-0.2mm}
\end{algorithm}


\subsection{Experiments on Face Recognition Datasets}

In this study, we employ CASIA\cite{yi2014learning} as our training data in order to conduct extensive comparisons with other softmax-based losses on face recognition datasets. For testing, in addition to the most widely used LFW\cite{huang2008labeled} , CFP-FP\cite{sengupta2016frontal}, CFP-FF\cite{sengupta2016frontal}, AgeDB\cite{Moschoglou2017AgeDB} datasets, we also report the performances on the recent datasets (e.g. CALFW\cite{zheng2017cross} and CPLFW\cite{zheng2018cross}) with large pose and age variations. 

The training settings of ArcFace follow the original ArcFace\cite{deng2019arcface} paper, and can be summarized as follows: all data are normalized and cropped to $112\times112$ according to the five facial points predicted by RetinaFace.
The widely used CNN architecture ResNet50 without the bottleneck structure is used as backbone. 
Following the original paper\cite{deng2019arcface}, 
we explore the BN-Dropout-FC-BN structure to obtain the final 512-D embedding feature after the last convolutional layer. 
The feature scale $s$ and the angular margin $m$ are set to 64 and 0.5 for ArcFace loss, respectively.
 
We employ the SGD optimizer and set the momentum to 0.9 and weight decay to 5e-4. The learning rate starts from 0.1 and decreases polynomially to 0 at the 20th epoch. 
We set the batch size to 256 and train the model on a single NVIDIA Tesla V100 (32GB) GPU. 
All experiments in this section are implemented with the open source codes from insightface\footnote{\href{https://github.com/deepinsight/insightface}{https://github.com/deepinsight/insightface}}. 
For comparison with our MadaCos, we only replace the ArcFace module with MadaCos module, where $\rho$ is set to 0.02 according to the main paper, and other settings are kept the same as ArcFace. As shown in Table.\ref{table:losscompare}, MadaCos outperforms re-implemented ArcFace$^r$, AdaFace$^r$, and ElasticFace$^r$ on most additional test sets, highlighting the strong generalizability of our approach.

\begin{table}[t!]
\scriptsize
\centering
\renewcommand\arraystretch{1.2}
\setlength{\tabcolsep}{4pt}
    \begin{tabular}{c|c|c|c|c|c|c}
    \hline
    Loss Functions   & LFW & CFP-FP & AgeDB & CALFW & CFP-FF & CPLFW \\
    \hline
    CosFace\cite{wang2018cosface} & 99.33 & - & - & - & - & - \\
    ArcFace\cite{deng2019arcface} & 99.53 & 95.56 & {\bf 95.15} & - & - & - \\
    ArcFace$^r$\cite{deng2019arcface} & 99.47 & 95.59 & 94.52 & 93.57 & 99.43 & 89.10 \\
    AdaFace\cite{kim2022adaface}$^r$ & 99.45 & {\bf 96.87} & 94.71 & 93.65 & 99.41 & 89.90 \\
    ElasticFace-Arc\cite{boutros2022elasticface}$^r$ & 99.38 & 96.39 & 94.78 & 93.33 & 99.39 & 89.38 \\
    ElasticFace-Cos\cite{boutros2022elasticface}$^r$ & 99.40 & 96.67 & 94.48 & 93.68 & 99.41 & 90.08 \\
    MadaCos & {\bf 99.57} & 96.51 & 95.12 & {\bf 93.90} & {\bf 99.59} & {\bf 90.20} \\
    \hline
    \end{tabular}
    \caption{Verification results ($\%$) of different loss functions. Methods superscripted with $^r$ are our re-implementations.}
    \label{table:losscompare}
\end{table}

\section{Ablation Studies}

\subsection{Different Parameters for MadaCos Loss}

To verify the robustness of our MadaCos loss in landmark datasets, we further adjust $\rho$ from 0.1 to 0.5 to train the global descriptors for 50 epochs, and the results are summarized in Table.\ref{table:madacos} an extended version of Table.\ref{table3} of the main paper. 
As $\rho$ increases, the mAP performance decreases, but it still surpasses the results of ArcFace loss. 

This proves that MadaCos does not require careful training tricks, and a smaller hyperparameter $\rho$ can better optimize the whole training process. 

\begin{table}
    \footnotesize
	\centering
	\renewcommand\arraystretch{1.2}
	\setlength{\tabcolsep}{7.2pt}
	\begin{tabular}{cccccc}
		\hline
		\multirow{2}{*}{Loss} &
		\multirow{2}{*}{$\rho$} &
		\multicolumn{2}{c}{Medium} &
		\multicolumn{2}{c}{Hard} \\
		\cline{3-6}
		& & $\mathcal{R}$Oxf & $\mathcal{R}$Par & 
		$\mathcal{R}$Oxf & $\mathcal{R}$Par\\
		\hline
		ArcFace\cite{deng2019arcface} & - & 78.76 & 88.59 & 58.53 & 75.97 \\
		\hline
		\multirow{5}{*}{MadaCos} & 0.1 & {\bf 80.56} & 89.86 & {\bf 61.25} & {\bf 78.57} \\
		 & 0.2 & 80.55 & 89.76 & 60.99 & 78.21 \\
		 & 0.3 & 79.82 & {\bf 89.98} & 60.72 & 78.44 \\
		 & 0.4 & 77.96 & 89.70 & 57.02 & 77.62 \\
		 & 0.5 & 78.96 & 88.99 & 59.79 & 76.48 \\
		\hline
	\end{tabular}
	\caption{Ablation study on different $\rho$ in MadaCos Loss.}
	\label{table:madacos}
\end{table}

\subsection{Triplet Loss Analysis}
In Table.\ref{triplet}, we provide more ablation experimental results about different $Q$ and $\mu$ in triplet loss. As $Q$ decreases, the model performance increases, while $\mu$ increases, which also slightly improves the performance. This proves that hard negatives are a critical key to feature learning, and hard negative sample is even more effective when used with MadaCos loss.

\begin{table}
    \small
	\centering
	\centering
    \setlength{\tabcolsep}{3.5mm}
	\begin{tabular}{cccccc}
		\hline
		\multirow{2}{*}{$Q$} &
		\multirow{2}{*}{$\mu$} &
		\multicolumn{2}{c}{Medium} &
		\multicolumn{2}{c}{Hard} \\
		\cline{3-6}
		& & $\mathcal{R}$Oxf & $\mathcal{R}$Par & 
		$\mathcal{R}$Oxf & $\mathcal{R}$Par\\
		\hline
		2 & 0.1 & 82.99 & 91.71 & 66.03 & 82.20 \\
		\hline
		6 & 0.1 & 82.42 & 91.57 & 65.06 & 81.69 \\
		\hline
		10 & 0.1 & 82.48 & 91.04 & 64.83 & 80.70 \\
		\hline
		6 & 0.2 & 82.91 & 91.52 & 66.07 & 81.82 \\
		\hline
	\end{tabular}
        \caption{Ablation study on different $Q$ and $\mu$ in triplet loss.}
	\label{triplet}
\end{table}

\subsection{Train epoch Analysis.} In Table.\ref{table2}, we start the next stage of training at the 50$th$ epoch because the first stage has converged. Additionally, in the 2048-dimension comparison group, we train the model for only 25 epochs. We fine-tune the global features starting from the 20$th$ epoch and continue for 5 epochs. In Table.\ref{epoch}, the model's performance improves with more epochs of training before the first stage of training converges. The two-stage training strategy significantly improves performance when the global features are fine-tuned on a better model.

\begin{table}
    \small
	\centering
	\vspace{0.3cm}
	\centering
	\setlength{\abovecaptionskip}{0cm}
    \setlength{\belowcaptionskip}{-1cm}
    \setlength{\tabcolsep}{2.2mm}
	\begin{tabular}{cccccc}
		\hline
        $E$ & $T$ & $\mathcal{R}$Oxf-M & $\mathcal{R}$Par-M & $\mathcal{R}$Oxf-H & $\mathcal{R}$Par-H \\
		\hline
            20 & 20 & 80.10 & 88.68 & 59.99 & 76.02 \\
 		20 & 25 & 82.51 & 89.64 & 63.59 & 78.06 \\
            13 & 13 & 78.38 & 88.54 & 57.63 & 76.41 \\
		13 & 25 & 80.54 & 88.97 & 61.35 & 76.51 \\
		\hline
	\end{tabular}
        \caption{Ablation study on different $E$ and $T$ in R50-CFCD with 2048 dimension. When $E=T$, we just train the first stage model.}
	\label{epoch}
\end{table}

\section{Comparison with Different Methods}
\subsection{Evaluation for the GLDv2 retrieval task.}
To evaluate our model, we report large-scale instance-level retrieval on the Google Landmarks dataset v2 (GLDv2-clean), using the latest ground-truth version (2.1). GLDv2 retrieval task has 1129 queries (379 validation and 750 testing) and 762k database images, with performance measured using mAP@100. In the same comparison group, our model's global retrieval performances show significant improvement, with an absolute gain of 6.8$\%$ for R101-DELG and 2.4$\%$ for R101-DOLG in Table.\ref{gldv2}.

\begin{table}
    \small
    \vspace{-0.0cm}
	\centering
	\setlength{\abovecaptionskip}{0cm}
    \setlength{\belowcaptionskip}{-0.3cm}
    \setlength{\tabcolsep}{1.2mm}
	\begin{tabular}{ccccc}
		\hline
        Methods & Testing & Validation & Scale & Dimension \\
		\hline
  		R101+ArcFace & 25.57 & 23.30 & - & - \\
            R101-DELG & 26.80 & - & 3 & 2048 \\
            R101-DOLG$^r$ & 31.08 & 27.97 & 5 & 512 \\
  		R101-CFCD & $\mathbf{33.60}$ & 30.85 & 3 & 2048 \\
		R101-CFCD & 33.51 & $\mathbf{30.93}$ & 5 & 512 \\
		\hline
	\end{tabular}
         \caption{Experimental results on GLDv2 dataset for retrieval task. Methods superscripted with $^r$ are our re-implementations.}
	\label{gldv2}
\end{table}

\begin{table}[t!]
\scriptsize
\centering
\renewcommand\arraystretch{1.2}
\setlength{\tabcolsep}{4pt}
    \begin{tabular}{ccccc}
    \hline
    Method   & \begin{tabular}[c]{@{}c@{}} Multi- \\Scale \end{tabular} & \begin{tabular}[c]{@{}c@{}} Global \\ Latency (ms) \end{tabular} & \begin{tabular}[c]{@{}c@{}} Matching \\ Latency (ms)\end{tabular} & \begin{tabular}[c]{@{}c@{}} Retrival \\ Latency (s)\end{tabular} \\
    \hline
    R50-CVNet-Rerank-400 & 3 & 25.78 & 43.63 & 1620.67  \\
    R50-CFCD & 5 & 35.12 & 0 & 889.12  \\
    \hline
    \end{tabular}
    \caption{Retrival Latency. ``Global Latency'' means the average extraction latency for multi-scale input with global model. ``Matching Latency'' means the average matching latency for CVNet-Rerank network. ``Retrival Latency'' means the total latency to preform retrieval on $\mathcal{R}$Oxf dataset. }
    \label{table:Latencycompare}
\end{table}

\subsection{Extraction Latency and Memory Footprint.}
In Table.\ref{table:Latencycompare}, we also conduct a study with $\mathcal{R}$Oxf dataset from latency perspective on NVIDIA Tesla V100 and Intel(R) Xeon(R) Platinum 8255C CPU. The latency of the state-of-the-art two-stage retrieval method CVNet\cite{lee2022cvnet} is nearly twice that of ours due to local re-ranking.
This means our CFCD is more practical to real-world applications as it achieves the approximate performance of two-stage methods while retaining the efficiency of single-stage methods. 



\end{document}